\documentclass[a4paper,conference]{IEEEtran}
\usepackage{cite}

\usepackage{graphicx}
\usepackage{amsmath}
\usepackage{mathtools}
\usepackage{amssymb}
\usepackage{algorithmic}

\usepackage{array}
\usepackage{fixltx2e}
\usepackage{dblfloatfix}

\usepackage{url}
\usepackage{float}

\usepackage{xcolor}
\newcommand{\etal}{\emph{et al. }}

\begin{document}
%
\title{Modeling Extent-of-Texture Information for\\ Ground Terrain Recognition}

\author{\IEEEauthorblockN{Shuvozit Ghose}
\IEEEauthorblockA{Institute of Engineering and \\Management, Kolkata, India\\
shuvozit.ghose@gmail.com}
\and
\IEEEauthorblockN{Pinaki Nath Chowdhury}
\IEEEauthorblockA{Indian Statistical Institute\\
Kolkata, India\\
contact@pinakinathc.me}
\and
\IEEEauthorblockN{Partha Pratim Roy}
\IEEEauthorblockA{Indian Institute of Technology\\
Roorkee, India\\
proy.fcs@iitr.ac.in}
\and
\IEEEauthorblockN{Umapada Pal}
\IEEEauthorblockA{Indian Statistical Institute\\
Kolkata, India\\
umapada@isical.ac.in}}


%


\maketitle

\begin{abstract}

Ground Terrain Recognition is a difficult task as the context information varies significantly over the regions of a ground terrain image. In this paper, we propose a novel approach towards ground-terrain recognition via modeling the Extent-of-Texture information to establish a balance between the order-less texture component and ordered-spatial information locally. At first, the proposed method uses a CNN backbone feature extractor network to capture meaningful information of ground terrain images, and model the extent of texture and shape information locally. Then, it encodes order-less texture information and ordered shape information in a patch-wise manner, and utilizes an intra-domain message passing mechanism to make every patch aware of each other for rich feature learning. Next, the model combines the extent of texture information with the encoded texture information and the extent of shape information with the encoded shape information patch-wise and then exploit Extent of texture (EoT) Guided Inter-domain Message passing module for sharing knowledge about the opposite domain to balance out the order-less texture information with ordered shape information. Finally, Bilinear model outputs a pairwise correlation between the order-less texture information and ordered shape information, and classifier classifies the ground terrain image efficiently. The experimental results indicate the superior performance of the proposed model\footnote[1]{The source code of the proposed system is publicly available at https://github.com/ShuvozitGhose/Ground-Terrain-EoT} over existing state-of-the-art techniques on DTD, MINC and GTOS-mobile datasets.

\end{abstract}


%
\IEEEpeerreviewmaketitle

\section{Introduction}
Ground terrain recognition is a popular area of research in the context of computer vision because of its widespread applications in robotics and automatic vehicular control \cite{angelova2007fast, manduchi2005obstacle, wolf2005autonomous, hebert2003terrain}. In the field of autonomous driving \cite{wolf2005autonomous}, ground terrain classification is very important because certain types of terrain may negatively affect the movement of a robot. Similarly, the knowledge of surrounded terrain information may help a robot to modify the course of its action during autonomous navigation \cite{manduchi2005obstacle, dahlkamp2006self}. The goal of ground terrain recognition is quite similar to that of object recognition, but various factors have made the ground terrain recognition quite a challenging task. Firstly, real-world ground terrain images usually have highly complicated terrain surfaces and may not have any obvious feature or edge points. Moreover, the terrain surface may be as highly complex as the surface of Earth. Secondly, many class boundaries of the ground terrain images are ambiguous. For example, the class \textit{``leaves"}  is similar to \textit{``grass"}, whereas the grass images contain a few leaves. Similarly, \textit{``asphalt"} class is similar to \textit{``stone-asphalt"} which is an aggregate mixture of stone and asphalt. Finally, the context information varies significantly over the regions of a ground terrain image, like some local regions possess significant texture information, while shape information is more dominant at some other parts.

Traditionally, ground terrain images are filtered with a set of handcrafted filter banks \cite{de1997multiresolution,cula2001compact,leung2001representing,konishi2000statistical}, followed by grouping outputs into bag-of-words or texton histograms for the purpose of ground terrain recognition.
\begin{figure}
  \centering
  \includegraphics[width=1\linewidth]{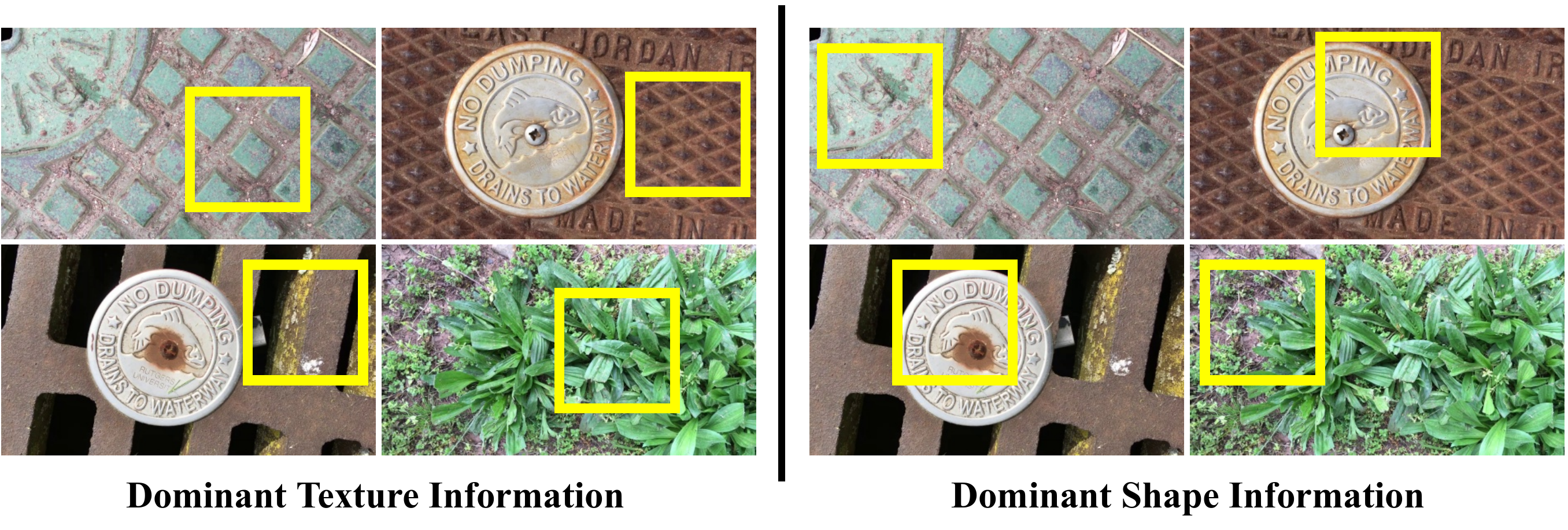}
  \caption{Examples of real-world ground terrain image. Left image shows the texture dominant local regions (yellow box), while the right shows shape dominant local regions.}
  \label{figure:2}
\end{figure}
Later, Cimpoi \etal \cite{cimpoi2016deep} developed deep filter banks based on the convolution layers of a deep model for texture recognition, description, and segmentation. A notable contribution of their work was the introduction of FV-CNN that replaced the handcrafted filter banks with pre-trained convolutional layers \cite{ghose2020fractional, bhunia2019novel, kishore2019user, banerjee2018local} for the feature extraction. Next, Zhang \etal \cite{zhang2017deep} introduced a Deep Texture Encoding Network with an encoding layer integrated on top of the convolution layers that could encode the texture information of the ground terrain and material surface images into order-less texture representation. This order-less representation was later used by the classifier for ground terrain and material recognition. The main drawback of their work was that they ignored shape information of the ground terrain and material surface images. To overcome this issue, Xue \etal \cite{xue2018deep} further presented Deep Encoding Pooling Network which integrated order-less texture details and global spatial information for the task of ground terrain recognition. But, from the Figure \ref{figure:2}, we can see that most real-world ground terrain images show wide variations in texture and shape information at different local regions in an image. While bounding boxes of the left image of Figure \ref{figure:2} show the texture dominant local regions, the bounding boxes of the right image of Figure \ref{figure:2} show the shape dominant local region. Thus, the classification of such realistic ground terrain images requires a more local level modeling of texture and shape information. For this reason, we propose a novel approach towards ground-terrain recognition via modeling the extent of texture information to establish a balance between the order-less texture and ordered-spatial information locally. We first use a CNN backbone feature extractor network to capture the meaningful information of the ground terrain image. Then, we model the extent of texture and shape information locally. Unlike \cite{zhang2016deep,xue2018deep}, we encode the order-less texture information and ordered-spatial information patch-wise. Next, we utilize Intra-domain Message passing mechanism to make every patch aware of each other for rich feature learning. Subsequently, we combine extent of texture information with encoded texture information, and the extent of shape information with the encoded shape information patch-wise to establish a more local level balance of the texture and shape information. Furthermore, we exploit Extent-of-Texture Guided Inter-domain Message passing module, for sharing knowledge about the other domain to balance out the order-less texture information with ordered shape information patch-wise. Next, we aggregate all the patches to get the global order-less texture and ordered shape information of the ground terrain image. Furthermore, a Bilinear model captures pairwise correlation between the order-less texture and ordered shape information. Finally, a classifier is used to classify the ground terrain image. The contribution of this paper is as follows:
\begin{itemize}
\item We propose a novel approach towards ground-terrain recognition by modeling the extent of texture information to establish a balance between the order-less texture and ordered-spatial information locally.

\item We introduce Intra-domain Message passing mechanism in the context of ground terrain recognition to make every local region aware of each other for rich feature learning.

\item We present Extent of texture Guided (EoT) Inter-domain Message passing
module in the context of ground terrain for sharing knowledge about the other domain to balance out the order-less texture information with ordered shape
information locally.

\item Our approach shows a superior classification accuracy on DTD, MINC and GTOS-mobile
datasets as compared to the previous state-of-the-arts methods.
\end{itemize}

The remaining of this paper is organized as follows. In section \ref{related}, we discuss some relevant works in the field of ground terrain recognition and graph convolutional neural networks. The proposed framework is detailed in Section \ref{section:proposed_method}. Section \ref{experiment} describes the datasets, implementation details, and experimental results. Section \ref{conclusion} concludes the paper.

\section{Related Works} \label{related}
Terrain Recognition is a well-known problem for decades in pattern recognition and computer vision community due to its crucial applications in the field of robotics. Earlier, the classical models were developed on geometric features, and Curvature-Based Approaches were exploited extensively in the context of terrain recognition. Goldgof \etal \cite{goldgof1989curvature} used a Gaussian and mean curvature profile for extracting special points on the terrain, and compared these specials points with the points of maximum and minimum curvature to recognize the particular regions of the terrain. Instead of using single geometrical feature, Yu and Yuan \cite{yu2009terrain} exploited multi-features including geometrical feature and color feature to classify terrain from ladar data for autonomous navigation. Based on the lighting direction of texture features, Chantler \etal \cite{chantler2002estimating} developed a probabilistic model which was robust to lighting direction and could classify the texture samples by comparing the likelihoods of each candidate with their estimated lighting. Further, Andreas \etal \cite{penirschke2002illuminant} exploited Lissajous’s ellipses to develop a classifier that could classify surface textures images under unknown illumination tilt angles. Manduchi \etal \cite{manduchi2005obstacle} introduced an obstacle detection technique based on stereo range measurements and proposed a color-based classification system to label the detected obstacles according to a set of terrain classes. One of the key features of this method was that it did not rely on typical structural assumption on the scene such as the presence of a visible ground plane for terrain classification. Cula and Dana \cite{cula2001compact} constructed a compact representation with the help of bidirectional texture function which captures the underlying statistical distribution of features in the image texture as well as the variations in this distribution with viewing and illumination direction. Thereafter, this compact representation was used by texture classifier, to classify texture images of unknown viewing and illumination direction efficiently. Leung and Malik \cite{leung2001representing} developed 3D textons on the basis of the textural appearance of material surfaces. Based on local geometric and photometric properties of the tiny surface patches of the material, the main idea was to construct a 3D texture vocabulary. Finally, a unified model was proposed to represent and recognize visual appearance of materials from the 3D textons vocabulary. Based on the Bidirectional Feature Histograms, Cula and Dana \cite{cula20043d} designed a 3D texture recognition method which employed the Bidirectional Feature Histograms as the surface model. The Bidirectional Feature Histograms captured the variation of the underlying statistical distribution of local structural image features, as the viewing and illumination conditions were changed; and classified surfaces based on a single texture image of unknown imaging parameter. Varma and Zisserman \cite{varma2005statistical} presented a statistical approach for texture classification from single images under unknown viewpoint and illumination. In this method, texture was modelled by the joint probability distribution of filter responses and was represented by the frequency histogram of filter response cluster centers. Bhunia \etal \cite{bhunia2019texture} used a generative approach towards texture retrieval. 

Later, Cimpoi \etal \cite{cimpoi2016deep} developed deep filter banks based on the convolution layers of a deep model for texture recognition, description, and segmentation. Instead of focusing on texture instance and material category, they proposed a human-interpretable vocabulary of texture attributes to describe common texture patterns. One of the key contributions of this work was the application of deep features to image regions and transferred features from one domain to another.
Zhang \etal \cite{zhang2018long} proposed deep encoder-decoder model with near-to-far learning strategy for the purpose of terrain segmentation. Next, Zhang \etal \cite{zhang2017deep} introduced Deep Texture Encoding Network with an encoding layer integrated on top of the convolution layers that could encode the texture image into order-less texture representation. This order-less representation was later used by a classifier for texture and material recognition.  Xue \etal \cite{xue2018deep} presented Deep Encoding Pooling Network which integrated order-less texture details with local spatial information for the task of ground terrain recognition. The framework learned a parametric distribution in feature space in a fully supervised manner and gave the distance relationship among classes to implicitly represent ambiguous class boundaries.

On the other hand, Graph based networks have gained popularity in many fields like machine translation \cite{cheng2016translation}, text recognition \cite{shi2015text} and learning sentence representation \cite{lin2017sentence}. Though convolution Neural Networks (CNN) have an amazing capability of feature extraction, their applications are limited to fixed grid-like structure. On the contrary, graph convolutional networks provide a simple alternative of feature extraction for the arbitrarily structures. The graph based feature extraction was first presented by Frasconi \etal \cite{frasconi1998graph} and Sperduti \etal \cite{sperduti1997graph}. They used recursive neural network on directed acylic graphs for the purpose of feature extraction. Later, Gori \etal \cite{gori2005graph} and Scarcelli \etal \cite{scarselli2009graph} proposed Neural Networks which extended the idea of recursive networks to both cyclic and acyclic types of graph structure. Recently, Velickovic \etal \cite{velickovic2017graphAN} proposed Graph Attention network which introduced the concept of attention mechanism in the context of Graph neural networks. Graph neural networks outputs a hidden representation for each node by attending to its neighbourhood nodes in coherence with a self-attention strategy. In the section \ref{GAT_layer}, we have exploited a graph attention network to facilitate every patch to attend to every other patch for rich feature learning.

\section{Proposed Framework}\label{section:proposed_method}

\subsection{Overview}

Contrary to the existing approaches \cite{zhang2016deep,xue2018deep} that neglect modeling of Extent-of-Texture (EoT) information, we propose a novel approach towards ground-terrain recognition by modeling EoT information to establish a balance between order-less texture component and ordered-spatial information. Our key observation is that the extent of texture information varies significantly over the regions of an image, such as some local regions possess significant texture information; while shape information is more dominant at some other parts. Therefore, we follow a patch based feature extraction approach in order to balance between the Texture ($\mathcal{T}$) and Shape ($\mathcal{S}$) domain locally. Overall, our framework could be grouped into five steps: (i) We introduce a novel way of modeling the EoT information. (ii) Off-the-shelf texture and shape feature extractors are employed to obtain patch-wise feature representation in each domain - $\mathcal{T}$ and $\mathcal{S}$. (iii) Intra-domain message passing mechanism is used to make every patch feature aware of the each other for rich feature learning. (iv) Thereafter, the patch feature from both $\mathcal{T}$ and $\mathcal{S}$ domains are combined guided by the EoT information. (v) Finally, we aggregate the patch features, followed by a bilinear operation to fuse two domains followed by two fully-connected layers for final classification.

\begin{figure*}[!hbtp]
  \includegraphics[width=1\linewidth]{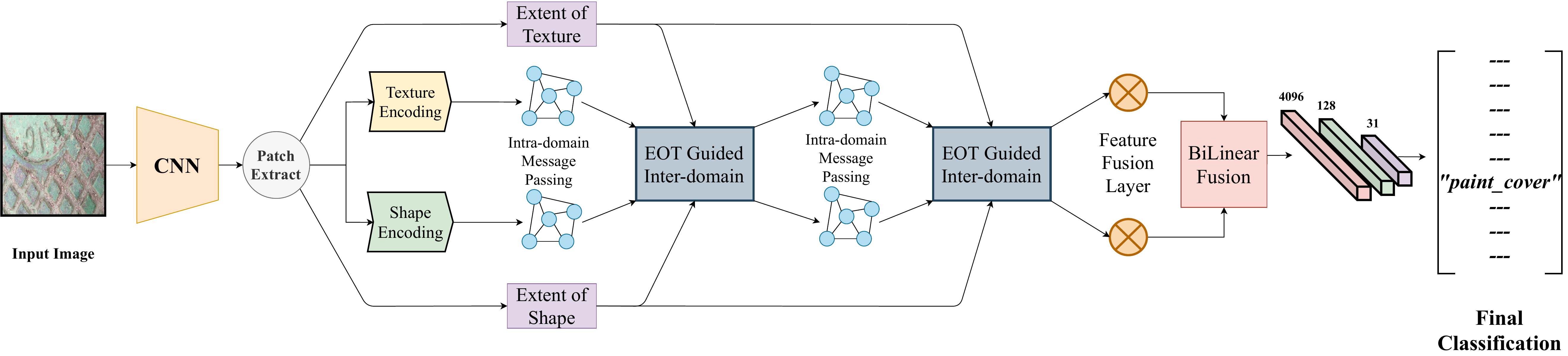}
  \caption{Our proposed Method. The CNN backbone feature extractor network reduces input to $Z \in 
  \mathbb{R}^{8 \times 8 \times 512}$. The patch extraction generates $36$ patches from $Z$. These patches are later used in Extent of texture modeling, texture encoding, shape encoding and Extent of shape modeling as shown in the diagram. The texture and shape encoding are passed through Intra-domain Message Passing and merged with Extent of texture and shape in the EoT Guided Inter-domain Message Passing module. The output texture and shape patches are combined in the feature fusion layers and then passed through Bilinear fusion layer. Finally, classifier is used.}
  \label{figure:1}

\end{figure*}

\subsection{Modeling Extent-of-Texture (EoT) Information}\label{extentTexture}
Extent-of-Texture (EoT) modeling is the primary step in our architecture. Let the ground terrain image be $\mathbf{I} \in \mathbb{R}^{H \times W \times 3}$, where H and W are the height and width of the ground terrain image. To capture the order-less texture information and ordered-spatial information effectively, we have used a backbone CNN feature extractor network $G(\cdot)$ that takes $\mathbf{I}$ as an input and outputs latent feature representation $\mathbf{Z}$. Thus,
\begin{equation}
\mathbf{Z}= G(\mathbf{I} ;\theta_{G})
\end{equation}
where $\mathbf{Z} \in  \mathbb{R}^{8 \times 8 \times 512}$ and $\theta_{G}$ is the parameter of the network. In general, $\mathbf{Z}$ is the latent feature representation of the local order-less texture information and ordered-spatial information. We adapt modified ResNet-18 architecture as CNN feature extractor network with rectified non-linearity (ReLU) activation after each layer. To model Extent-of-Texture (EoT) information locally, we perform patch-extraction on $\mathbf{Z}$ using a sliding window mechanism where the window size and stride is chosen as $(3 \times 3)$ and $1$ respectively. The patch-extraction operation generates $\mathbf{\psi} =\begin{Bmatrix}
\psi_{1}, \psi_{2},\psi_{3}.......\psi_{k}
\end{Bmatrix}$
patches, where $\mathbf{\psi_{i}} \in \mathbb{R}^{3 \times 3 \times 512}$ and $k$ is the number of patches. In our model, the value of $k$ is $36$. Next, an average pooling operation with kernel size $(3 \times 3)$ is performed on $\psi$ which results in $\mathbf{\psi^{*}} =\begin{Bmatrix}
\psi_{1}^{*}, \psi_{2}^{*},\psi_{3}^{*}.......\psi_{k}^{*}
\end{Bmatrix}$
patches, where $\mathbf{\psi_{i}^{*}} \in \mathbb{R}^{1 \times 1 \times 512}$. Let $\mathbf{X} =\begin{Bmatrix}
x_{1}, x_{2},x_{3}.......x_{k}, 
\end{Bmatrix}$, where $x_{i}$ denotes the central region of the  $\mathbf{\psi_{i}}$ patch e.g. $x_{i} = \psi_{i}[2;2;:]$ and ${x_{i}} \in \mathbb{R}^{1 \times 1 \times 512}$. The cosine similarity between $\mathbf{\psi^{*}}$ and $\mathbf{X}$ describes the order-less texture information $\mathcal{T}$, where $\mathcal{T} = \begin{Bmatrix}
\mathcal{T}_{1}, \mathcal{T}_{2},\mathcal{T}_{3}.......\mathcal{T}_{k} 
\end{Bmatrix}$ and $\mathcal{T}_{i}$ denotes the order-less texture information of the $i^{th}$ patch. Therefore,
\begin{equation}
        \psi_{i}^{*}= AvgPool(\psi_{i}, 3)
\end{equation}
\begin{equation}
    \begin{gathered}
        \mathcal{T}_{i} = \frac{\psi_{i}^{*} \cdot x_{i}}{||\psi_{i}^{*}||_{2} \ ||x_{i}||_{2}}
    \end{gathered}
\end{equation}
\begin{equation} \label{equa4}
    \begin{gathered}
        \mathcal{T}_{i} = \frac{\mathcal{T}_{i} - \mathcal{T}_{min}}{\mathcal{T}_{max} - \mathcal{T}_{min}}
    \end{gathered}
\end{equation}
Here, $||\cdot||_{2}$ represents L2-norm, $\mathcal{T}_{min}$ and $\mathcal{T}_{max}$ are the fixed value of $0.5$ and $0.9$ respectively. In our experiment, we have observed that the cosine distance between $\psi^{*}$ and $x_{i}$ varies between $0.5$ and $0.9$. We perform a normalization operation on $\mathcal{T}$ using equation \ref{equa4} for readjusting the range from $0$ to $1$. In this context, a high value of $\mathcal{T}$ indicates the presence of greater extent of the order-less texture information , whereas a small value of $\mathcal{T}$ represents higher shape information. Let the ordered shape information $\mathcal{S}$, where $\mathcal{S} = \begin{Bmatrix}
\mathcal{S}_{1}, \mathcal{S}_{2},\mathcal{S}_{3}.......\mathcal{S}_{k} 
\end{Bmatrix}$ and $\mathcal{S}_{i}$ denotes the ordered-spatial information of the $i^{th}$ patch. Then,
\begin{equation}
        \mathcal{S}_{i}= 1-\mathcal{T}_{i}
\end{equation}

\subsection{Texture and Shape Encoding Module}
In our architecture, we have used an off-the-shelf texture encoding layer proposed by Zhang \etal \cite{zhang2016deep} for texture encoding. The texture encoding layer integrates the entire dictionary learning and visual encoding pipeline to provide an order-less representation for texture modeling. For each $\psi_{i} \ \in \psi$, let $\mathbf{\delta} = \{ \delta_{1}, \delta_{2}, \dots, \delta_{m} \}$ be M visual descriptors and $\mathbf{\Lambda} = \{ \lambda_{1}, \lambda_{2}, \dots \lambda_{n} \}$ be N learned codewords. We calculate the residual vectors $r_{i,j} = \delta_{i} - \lambda_{j}$ where, $i =  1,2, \dots m $ and $j =  1,2, \dots n $. The residual encoding corresponding to $\delta_{j}$ is calculated as:
\begin{equation}
    t_{j} = \sum_{i=1}^{M} w_{i,j} r_{i,j}
\end{equation}
\begin{equation}
\begin{split}
    \text{ where}, w_{i,j} = \frac{exp(-s_{j}||r_{i,j}||^{2})}{\sum_{l=1}^{N} exp(-s_{l}||r_{i,l}||^{2})}
\end{split}
\end{equation}
Here $ s_{1}, s_{2}, \dots s_{N} $ are learnable smoothing factors for each cluster center $\lambda_{j} \in \mathbf{\Lambda}$. Let $\mathbf{E} = \{ t_{1}, t_{2}, \dots t_{n} \}$ be the encoded order-less texture information, where $t_{i}$ having a dimension of $\mathbb{R}^{512 \times N}$ is fed to a fully connected layer $f_{c}:\mathbb{R}^{4096} \rightarrow \mathbb{R}^{F}$ to give $\mathbf{E_{t}} = \{ e_{t1}, e_{t2}, \dots e_{tn} \}$, where $e_{ti} \ \in \ \mathbb{R}^{F}$ and $e_{ti}$ represents the final encoded order-less texture information of the $i^{th}$ patch. Here, N is the number of learned codewords and F is the size of output of the fully connected layer. In our architecture, we have used $N=8$ and $F=64$.

For shape encoding, first we have performed average pooling on each $\psi_{i} \in \psi$ that converts $\mathbb{R}^{ 3 \times 3 \times 512} \rightarrow \mathbb{R}^{1 \times 1 \times 512}$. Next, a fully connected layer maps $f_{c}:\mathbb{R}^{512} \rightarrow \mathbb{R}^{F}$ to give $\mathbf{E_{s}} = \{ e_{s1}, e_{s2}, \dots e_{sn} \}$, where $e_{si} \ \in \ \mathbb{R}^{F}$ and  $e_{si}$ represents the final encoded ordered shape information of the $i^{th}$ patch.

\subsection{Intra-domain Message Passing} \label{GAT_layer}
We develop an Intra-domain Message Passing module to make every patch feature aware of each other for rich feature learning. A graph attention layer (GAT) \cite{velickovic2017graphAN} is employed to allow every patch to attend to every other patches. For this reason, we have designed two separate complete graphs, each having $k$ nodes, where the degree of each node is $k$. The $i^{th}$ node of the $E_{t}$ representational graph represents the $e_{ti}$ patch of  $E_{t}$, and the $i^{th}$ node of the $E_{s}$ representational graph represents the $e_{si}$ patch of $E_{s}$. So, graph $\mathbf{E_{t}} = \{ e_{t1}, e_{t2}, \dots e_{tk} \}$, where $e_{ti} \in \mathbb{R^{F}}$ and F is the number of features in each node (in our case, $F=64$). We compute the hidden representation for each $e_{ti}$ by attending to all $e_{tj}$ where $j= 1,2,\dots k$ using self-attention mechanism. To transform $e_{ti}$ to higher-level features, a shared linear transformation matrix $\mathbf{W} \in \mathbb{R}^{F \times F}$ is applied to each $e_{ti}$. This is followed by encompassing self attention over all nodes using a shared attention mechanism orchestrated by a vector $\mathbf{\vec{a}} \in \mathbb{R}^{2F}$. The final representation $e'_{ti}$ for each $e_{ti}$ is computed as follows:
\begin{equation}\label{eq_atten}
    \begin{gathered}
        e_{i,j} = LeakyReLU(\vec{\mathbf{a}}^{T} [\mathbf{W}e_{ti} || \mathbf{W}e_{tj}]) \\
    \end{gathered}
\end{equation}
\begin{equation}
        \alpha_{i,j} = softmax_{j}(e_{i,j}) = \frac{exp(e_{i,j})}{\sum_{l=1}^{k}exp(e_{i,l})} \
\end{equation}
\begin{equation}
\begin{gathered}
        e'_{ti} = \sigma(\sum_{j=1}^{k} \alpha_{i,j} \mathbf{W}e_{tj})
    \end{gathered}
\end{equation}
where $||$ represents concatenation of two vectors, $\sigma$ represents a Relu non-linear activation function and k is the number of nodes connected to $i$. We use multi-head attention to stabilise the self-attention mechanism in GAT layers. The GAT layer outputs $\mathbf{E'_{t}} = \{ e'_{t1}, e'_{t2}, \dots e'_{tk} \}$ for $\mathbf{E_{t}}$ representational graph. Similarly, we obtain $\mathbf{E'_{s}} = \{ e'_{s1}, e'_{s2}, \dots e'_{sk} \}$ for $\mathbf{E_{s}}$ representational graph.

\begin{figure*}[t]
  \centering
  \includegraphics[width=1\linewidth]{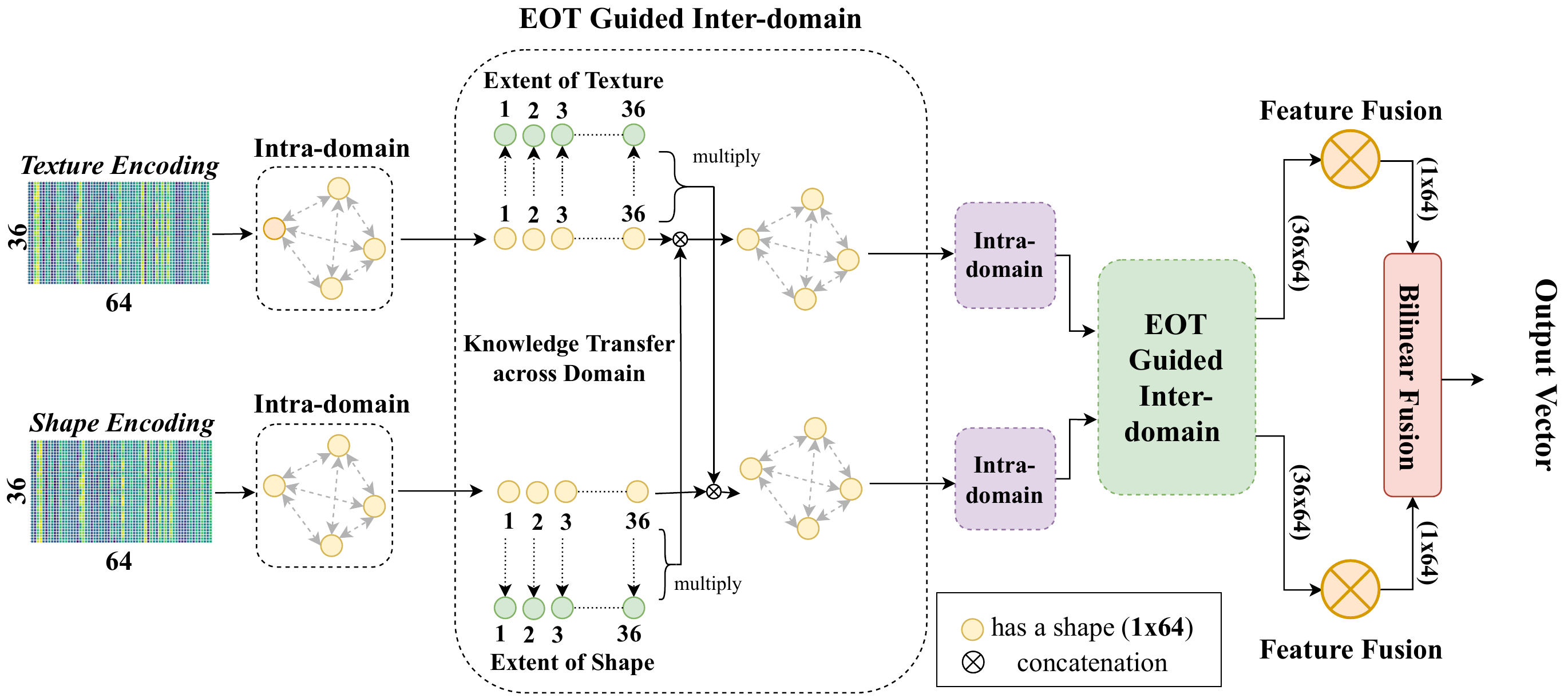}
  \caption{Architecture of the EoT Guided Inter-domain Message Passing. Feature map of the $36$ patches from Texture encoding and shape encoding layer, having shape $\mathbb{R}^{36 \times 64}$ is given to Intra-domain Message Passing module for information exchange among the patches. From the resulting $36$ features for texture and shape, a fused vector is calculated using weighted average where extent of texture/shape serves as the weights. This fused vector is concatenated with each of the $36$ feature vectors from the other domain followed by a fully connected layer to reduce dimensionality of each vector back to $64$. We further enrich the representation using a second Intra-domain and EoT Guided Inter-domain Message Passing module before fusing the $36$ vector of each domain in the Fusion layer followed by Bilinear fusion.}
  \label{figure:3}

\end{figure*}

\subsection{EoT Guided Inter-domain Message Passing}\label{CDKT_section}
In this section, first we have combined the extent of texture information $\mathcal{T}$ with $E'_{t}$ and the extent of shape information $\mathcal{S}$ with $E'_{s}$ to establish a local level balance of the order-less texture and ordered shape information. Later, the EoT Guided Inter-domain Message Passing module is used for sharing knowledge about the other domain to balance out the order-less texture information with ordered-spatial information as depicted in the Figure \ref{figure:3}. Mathematically,
\begin{equation}
    \begin{split}
        r_{t}^{'} = \sum_{j=1}^{k}\mathcal{T}_{j}e_{tj}^{'} 
    \end{split}
\end{equation}
\begin{equation}
    \begin{split}
        r_{s}^{'} = \sum_{j=1}^{k}\mathcal{S}_{j}e_{sj}^{'}
    \end{split}
\end{equation}
where $k$ is the number of patches, $r_{i}^{'t}$ and $r_{i}^{'s}$ are the representative vectors of texture and shape domains respectively. $\mathcal{T}_{j} \in \mathcal{T}$ and $\mathcal{S}_{j} \in \mathcal{S}$ are described in section \ref{extentTexture}, whereas $e_{tj}^{'} \in E'_{t}$ and $e_{sj}^{'} \in E'_{s}$ are described in section \ref{GAT_layer}. Let $\mathbf{W}_{t}$ and $\mathbf{W}_{s}$ are a pair of shared learnable weights, the EoT Guided Inter-domain Message Passing module generates
\begin{equation}
    \begin{split}
        \mathcal{T}_{i}^{'} = \mathbf{W}_{t} [e_{ti}^{'}||r_{s}^{'}] 
    \end{split}
\end{equation}
\begin{equation}
\begin{split}
        \mathcal{S}_{i}^{'} = \mathbf{W}_{s} [e_{si}^{'}||r_{t}^{'}]
    \end{split}
\end{equation}
Where, $\mathcal{T}_{i}^{'}$ and $\mathcal{S}_{i}^{'}$ are the balanced texture and shape information of the $i^{th}$ patch respectively. We further enrich this representation using a succession of Intra-domain Message Passing module and EoT Guided Inter-domain Message Passing module to derive  $\mathcal{T}^{''}=\{\mathcal{T}_{1}^{''}, \mathcal{T}_{2}^{''},\dots \mathcal{T}_{k}^{''}\}$ and $\mathcal{S}^{''}=\{\mathcal{S}_{1}^{''}, \mathcal{S}_{2}^{''},\dots \mathcal{S}_{k}^{''}\}$, where each $i^{th}$ patch having a dimension of $\mathbb{R}^{K \times F}$.

\subsection{Fusion Layer and Bilinear Model}
In the feature fusion layer, we have aggregated the all $\mathcal{T}_{i}^{''}$ and $\mathcal{S}_{i}^{''}$ patches for $i=1,2,\dots k$ to get global order-less texture representation $\mathcal{T}_{fuse}$ and ordered shape representation $\mathcal{S}_{fuse}$ respectively. Let $\mathcal{T}^{''}$ and $\mathcal{S}^{''}$ be two matrices of size $\mathbb{R}^{k \times F}$, where each column of $\mathcal{T}^{''}$  and $\mathcal{S}^{''}$ represents $\mathcal{T}_{i}^{''}$ and $\mathcal{S}_{i}^{''}$ patches respectively. Then,
\begin{equation}
    \begin{split}
        \mathcal{T}^{fuse} = \mathcal{V}_{t}^{T}\mathcal{T}^{''}
    \end{split}
\end{equation}
\begin{equation}
    \begin{split}
        \mathcal{S}^{fuse} = \mathcal{V}_{s}^{T}\mathcal{S}^{''} 
    \end{split}
\end{equation}
Where, $\mathcal{V}_{t}$ and $\mathcal{V}_{s}$ are learnable weight vectors, each having a size $\mathbb{R}^{k}$. Next, $\mathcal{T}_{fuse}$ and $\mathcal{S}_{fuse}$ are passed through the Bilinear fusion model\cite{xue2018deep} to balance the texture and shape information. Let $f_{out}$ be the final output of the Bilinear fusion layer. Then,
\begin{equation}
    \begin{split}
        f_{out} = \sum_{i=1}^{F}\sum_{j=1}^{F}\omega_{i,j}\mathcal{T}_{i}^{fuse}\mathcal{S}_{j}^{fuse}
    \end{split}
\end{equation}
where, $f_{out}\in\mathbb{R}^{4096}$ and $\omega_{i,j}$ is a learnable weight to balance the interaction between texture and shape information. Here, the $f_{out}$ captures a pairwise correlation between the order-less texture and ordered shape information. Finally, $f_{out}$ is passed through a sequence of two fully-connected layers to generate the classification result of the ground-terrain image as depicted in the Figure \ref{figure:1}.

\section{Experiment}\label{experiment}

\subsection{Datasets}
\textbf{GTOS-mobile} \cite{xue2018deep} dataset, which is a modified version of the original GTOS database \cite{zhang2016deep}, is used for evaluation of our approach. GTOS-mobile dataset consists of 31 classes captured using mobile phones consisting of 81 videos from which 6066 frames were used as a test set. We follow the train-test split as described in Xue \etal \cite{xue2018deep}. GTOS-mobile differs from the original GTOS dataset where the classes: dry grass, ice mud, mud-puddle, black ice and snow were removed. Further similar classes such as asphalt and metal are merged. Hence, unlike the original GTOS dataset consisting of 40 classes, the GTOS-mobile dataset has 31 classes as defined in Xue \etal \cite{xue2018deep}. The resolution of videos was maintained at $1920 \times 1080$ followed by resizing the shorter edge to $256$. Hence each image has a resolution of $455 \times 256$. 

\textbf{Describable Texture Dataset}(DTD) \cite{cimpoi2014dtd} consists of $47$ classes with each class consisting of $120$ instances. The images are collected from Google and Flickr using key attributes for each class and a list of joint attributes. Annotation of images were carried out using Amazon Mechanical Turk in several iterations.

\textbf{Materials in Context Database - 2500}(or MINC-2500) \cite{bell2015minc} is a subset of the original MINC dataset with $2500$ instances in every class/category where each image has been resized to $362 \times 362$.

\subsection{Implementation Details}
We have implemented the entire model in PyTorch \cite{paszke2017automatic} and executed the code on a server having Nvidia Titan X GPU with 12 GB of memory. L2 loss was employed as the objective function during experimentation. We have adapted a modified ResNet-18 architecture as CNN feature extractor network with rectified non-linearity (ReLU) activation after each layer. The full resolution input image is resized into different scales followed by cropping the center $256 \times 256$ regions because such a pre-processing step simulates observing the GTOS dataset at different distances. During training, we have initialized our CNN backbone feature extractor network using pre-trained ImageNet \cite{imagenet_cvpr09} weights. Following previous works in texture recognition \cite{xue2018deep}, we follow single-scale training and multi-scale training with identical data augmentation and training procedures. For single-scale training mechanism, an input image is resized into $286 \times 286$ and crop a region of size $256 \times 256$ from the center. Multi-scale training incorporates randomly resizing an input image into $256 \times 256$, $384 \times 384$, and $512 \times 512$ followed by cropping a region of $256 \times 256$ from the image center. The training data additionally undergoes a $50\%$ chance of horizontal flip with random changes in brightness, contrast and saturation. Images in both training and testing sets, finally undergo per-channel normalisation before being fed to the model. Training is performed for $30$ epochs with a batch size of 128. We have used stochastic gradient descent (SGD) optimizer with learning rate $0.01$, momentum $0.9$, decay rate of $0.0001$ while training our model in an end-to-end fashion.

\subsection{Baselines Methods}
\begin{table*}
	\centering
    \caption{Comparison of \textbf{Deep-TEN}, baseline \textbf{B1, B2, B3} and \textbf{B4} with the proposed methodology for single scale and multi scale training on GTOS-mobile \cite{xue2018deep} dataset using a pre-trained ResNet-18 module as the convolutional layer. Baseline B1 is similar to Deep Encoding Pooling Network (DEP) by Xue \etal \cite{xue2018deep}.}
    \begin{tabular}{c|c|c|c|c|c|c}
    \hline \hline
    & \textbf{Deep-TEN \cite{zhang2016deep}} & \textbf{B1 \cite{xue2018deep}} & \textbf{B2} & \textbf{B3} & \textbf{B4} & \textbf{Proposed Method} \\ \hline
    Single Scale & 74.22 & 76.07 & 77.81 & 78.55 & 78.93 & \textbf{80.39} \\
    Multi Scale & 76.12 & 82.18 & 83.78 & 84.31 & 84.36 & \textbf{85.71} \\ \hline
    \end{tabular}
    \label{tab:baseline}
\end{table*}

To justify each design choice and their contribution, we design four alternative baselines.

\textbf{Baseline-1 (B1):} Following the existing DEP \cite{xue2018deep} method, we feed the features from convolutional layers to the texture and shape encoding layers. Output from the encoding layers are then merged together using Bilinear model \cite{tenenbaum1997bilinear,lin2015bilinear}. 

\textbf{Baseline-2 (B2):} We construct a baseline where features from convolutional layers are segmented into patches using patch extraction and fed to both texture encoding and global average pooling. After passing through texture and shape encoding layers, the feature vectors of patches corresponding to both texture and shape are merged using the aforementioned Fusion layer to get a global representative vector for texture and shape respectively. These vectors are then combined using Bilinear models for eventual classification. 

\textbf{Baseline-3 (B3):} Here we apply Intra-domain Message Passing module using Graph Attention networks \cite{velickovic2017graphAN} to derive features from the texture and shape encoding layer. This is followed by Fusion and bilinear layer, similar to B2. 

\textbf{Baseline (B4):} We replace the Intra-domain Message Passing module in baseline B3 with the EoT Guided Inter-domain Message Passing module to enable a better balance between order-less texture component with ordered shape information. 

We compare the aforementioned baselines and Deep-TEN \cite{zhang2016deep} with our proposed methodology to examine the contribution of each design choice. We maintain an identical training and evaluation procedure where ResNet-18 generates a feature map of dimension $8 \times 8 \times 512$. Experiments on GTOS-mobile dataset are shown in Table \ref{tab:baseline} which demonstrates that the proposed methodology improves upon Deep-TEN \cite{zhang2016deep} by nearly $6\%$ and DEP \cite{xue2018deep} by nearly $4\%$.

\subsection{Performance Analysis}
From Table \ref{tab:baseline}, we observe that B1 has achieved a significant improvement of $1.85\%(6.06\%)$ from Deep-TEN \cite{zhang2016deep} for Single-scale (Multi-scale) setup by incorporating the spatial information instead of replying solely on texture encoding layer. Although B1 incorporated both texture and spatial information, it did not account for the local level variations of the order-less texture and ordered-spatial information. Using patch extraction in B2, we mitigate this limitation and further improve recognition performance over B1 by $1.74\%(1.6\%)$ for Single-scale (Multi-scale). While B2 was able to enhance recognition accuracy by taking a more local level approach through patch extraction, it did not correlate the patch information with each other to enrich features and hence, we observe an inferior performance as compared to B3. Using Intra-domain Message Passing to make every patch aware of each other via Graph-Attention layer resulted in an improvement  of $0.74\%(0.53\%)$ over B2. In B4, replacing graph attention layer in B3 by the EoT Guided Inter-domain Message Passing mechanism resulted in an improvement of $1.12\%(0.58\%)$ over B2, indicating the significance of information exchange across different domains by the EoT Guided Inter-domain Message Passing module. While both B3 and B4 show considerable improvements over B2, it can be observed from Table \ref{tab:baseline} that B4 performs slightly better than B3 by $0.38\%(0.05\%)$. This implies that sharing knowledge across texture and shape domains before fusing the local texture and shape features in the Fusion layer is slightly more beneficial as compared to exchanging knowledge among similar entities followed by merging texture and spatial information in the Fusion layer.

\subsection{Comparison on DTD and MINC datasets}
\begin{table}
	\centering
    \caption{Comparing Our method with several state-of-the-art methods on Describable Textures Dataset (DTD) and Materials in Context Database (MINC)}
	\begin{tabular}{c|c|c} \hline \hline
		Method & DTD \cite{cimpoi2014dtd} & MINC-2500 \cite{bell2015minc} \\ \hline
        FV-CNN \cite{cimpoi2015fvcnn} & 72.3 & 63.1 \\
        Deep-TEN \cite{zhang2016deep} & 69.6 & 80.4 \\
        DEP \cite{xue2018deep} & 73.2 & 82.0 \\
        \textbf{Proposed Method} & \textbf{75.7} & \textbf{85.3}  \\ \hline
	\end{tabular}
    \label{tab:comparison}
\end{table}

To ensure an equal comparison, we replace ResNet-18 with ResNet-50 and include a $(1 \times 1)$ convolutional layer to convert the number of output channels from $2048$ to $512$. Evaluation on Describable Textures Database (DTD) \cite{cimpoi2014dtd} and Materials in Context Database (MINC) \cite{bell2015minc} expresses the generalisability of the proposed method. From Table \ref{tab:comparison}, we can observe that for DTD (MINC-2500) dataset, the proposed method shows $2.5\%(3.3\%)$ improvement as compared to state-of-the-art methods. Additionally, a multi-scale training mechanism is likely to improve fine-grained visual recognition results for all methods as demonstrated by Lin \etal \cite{lin2015bilinear}. Although, we do not include a multi-scale training setup in our experimental section, one can expect enhancement of performance for both the proposed method as well as existing baselines by using multi-scale training.

\section{Conclusion} \label{conclusion}
In this paper, we have proposed a novel approach towards ground-terrain recognition via modeling the extent of texture information to establish a balance between the order-less
texture component and ordered-spatial information locally. The driving idea of our architecture is the modeling of context information locally. The proposed framework is simple and easy to implement. It is capable of detecting ground terrain in the real-world scenario. We demonstrate the effectiveness of our system by conducting experiments on publicly available ground terrain datasets.


\bibliographystyle{IEEEtran}
\bibliography{IEEEexample}

\begin{thebibliography}{10}
\providecommand{\url}[1]{#1}
\csname url@samestyle\endcsname
\providecommand{\newblock}{\relax}
\providecommand{\bibinfo}[2]{#2}
\providecommand{\BIBentrySTDinterwordspacing}{\spaceskip=0pt\relax}
\providecommand{\BIBentryALTinterwordstretchfactor}{4}
\providecommand{\BIBentryALTinterwordspacing}{\spaceskip=\fontdimen2\font plus
\BIBentryALTinterwordstretchfactor\fontdimen3\font minus
  \fontdimen4\font\relax}
\providecommand{\BIBforeignlanguage}[2]{{%
\expandafter\ifx\csname l@#1\endcsname\relax
\typeout{** WARNING: IEEEtran.bst: No hyphenation pattern has been}%
\typeout{** loaded for the language `#1'. Using the pattern for}%
\typeout{** the default language instead.}%
\else
\language=\csname l@#1\endcsname
\fi
#2}}
\providecommand{\BIBdecl}{\relax}
\BIBdecl

\bibitem{angelova2007fast}
A.~Angelova, L.~Matthies, D.~Helmick, and P.~Perona, ``Fast terrain
  classification using variable-length representation for autonomous
  navigation,'' in \emph{CVPR}, 2007.

\bibitem{manduchi2005obstacle}
R.~Manduchi, A.~Castano, A.~Talukder, and L.~Matthies, ``Obstacle detection and
  terrain classification for autonomous off-road navigation,'' \emph{Autonomous
  robots}, 2005.

\bibitem{wolf2005autonomous}
D.~F. Wolf, G.~S. Sukhatme, D.~Fox, and W.~Burgard, ``Autonomous terrain
  mapping and classification using hidden markov models,'' in \emph{ICRA},
  2005.

\bibitem{hebert2003terrain}
M.~Hebert and N.~Vandapel, ``Terrain classification techniques from ladar data
  for autonomous navigation,'' 2003.

\bibitem{dahlkamp2006self}
H.~Dahlkamp, A.~Kaehler, D.~Stavens, S.~Thrun, and G.~R. Bradski,
  ``Self-supervised monocular road detection in desert terrain,'' in
  \emph{Robotics: science and systems}, 2006.

\bibitem{de1997multiresolution}
J.~S. De~Bonet, ``Multiresolution sampling procedure for analysis and synthesis
  of texture images,'' in \emph{ACM Annual Conference on Computer graphics and
  interactive techniques}, 1997.

\bibitem{cula2001compact}
O.~G. Cula and K.~J. Dana, ``Compact representation of bidirectional texture
  functions,'' in \emph{CVPR}, 2001.

\bibitem{leung2001representing}
T.~Leung and J.~Malik, ``Representing and recognizing the visual appearance of
  materials using three-dimensional textons,'' \emph{IJCV}, 2001.

\bibitem{konishi2000statistical}
S.~Konishi and A.~L. Yuille, ``Statistical cues for domain specific image
  segmentation with performance analysis,'' in \emph{CVPR}, 2000.

\bibitem{cimpoi2016deep}
M.~Cimpoi, S.~Maji, I.~Kokkinos, and A.~Vedaldi, ``Deep filter banks for
  texture recognition, description, and segmentation,'' \emph{IJCV}, 2016.

\bibitem{ghose2020fractional}
S.~Ghose, A.~Das, A.~K. Bhunia, and P.~P. Roy, ``Fractional local neighborhood
  intensity pattern for image retrieval using genetic algorithm,''
  \emph{Multimedia Tools and Applications}, 2020.

\bibitem{bhunia2019novel}
A.~K. Bhunia, A.~Bhattacharyya, P.~Banerjee, P.~P. Roy, and S.~Murala, ``A
  novel feature descriptor for image retrieval by combining modified color
  histogram and diagonally symmetric co-occurrence texture pattern,''
  \emph{Pattern Analysis and Applications}, 2020.

\bibitem{kishore2019user}
P.~S.~R. Kishore, A.~K. Bhunia, S.~Ghose, and P.~P. Roy, ``User constrained
  thumbnail generation using adaptive convolutions,'' in \emph{ICASSP}, 2019.

\bibitem{banerjee2018local}
P.~Banerjee, A.~K. Bhunia, A.~Bhattacharyya, P.~P. Roy, and S.~Murala, ``Local
  neighborhood intensity pattern--a new texture feature descriptor for image
  retrieval,'' \emph{Expert Systems with Applications}, 2018.

\bibitem{zhang2017deep}
H.~Zhang, J.~Xue, and K.~Dana, ``Deep ten: Texture encoding network,'' in
  \emph{CVPR}, 2017.

\bibitem{xue2018deep}
J.~Xue, H.~Zhang, and K.~Dana, ``Deep texture manifold for ground terrain
  recognition,'' in \emph{CVPR}, 2018.

\bibitem{zhang2016deep}
H.~Zhang, J.~Xue, and K.~Dana, ``Deep ten: Texture encoding network,''
  \emph{arXiv preprint arXiv:1612.02844}, 2016.

\bibitem{goldgof1989curvature}
D.~B. Goldgof, T.~S. Huang, and H.~Lee, ``A curvature-based approach to terrain
  recognition,'' \emph{TPAMI}, 1989.

\bibitem{yu2009terrain}
C.-p. Yu and X.~Yuan, ``Terrain classification for autonomous navigation using
  ladar sensing,'' in \emph{International Conference on Information Science and
  Engineering}, 2009.

\bibitem{chantler2002estimating}
M.~J. Chantler, G.~McGunnigle, A.~Penirschke, and M.~Petrou, ``Estimating
  lighting direction and classifying textures.'' in \emph{BMVC}, 2002.

\bibitem{penirschke2002illuminant}
A.~Penirschke, M.~J. Chantler, and M.~Petrou, ``Illuminant rotation invariant
  classification of 3d surface textures using lissajous’s ellipses,'' in
  \emph{International Workshop on Texture Analysis and Synthesis}, 2002.

\bibitem{cula20043d}
O.~G. Cula and K.~J. Dana, ``3d texture recognition using bidirectional feature
  histograms,'' \emph{IJCV}, 2004.

\bibitem{varma2005statistical}
M.~Varma and A.~Zisserman, ``A statistical approach to texture classification
  from single images,'' \emph{IJCV}, 2005.

\bibitem{bhunia2019texture}
A.~K. Bhunia, S.~R.~K. Perla, P.~Mukherjee, A.~Das, and P.~P. Roy, ``Texture
  synthesis guided deep hashing for texture image retrieval,'' in \emph{WACV},
  2019.

\bibitem{zhang2018long}
W.~Zhang, Q.~Chen, W.~Zhang, and X.~He, ``Long-range terrain perception using
  convolutional neural networks,'' \emph{Neurocomputing}, 2018.

\bibitem{cheng2016translation}
C.~Jianpeng, D.~Li, , and L.~Mirella, ``Long short-term memory-networks for
  machine reading,'' \emph{arXiv preprint arXiv:1601.06733}, 2016.

\bibitem{shi2015text}
B.~Shi, X.~Bai, and C.~Yao, ``An end-to-end trainable neural network for
  image-based sequence recognition and its application to scene text
  recognition,'' \emph{TPAMI}, 2015.

\bibitem{lin2017sentence}
L.~Zhouhan, F.~Minwei, N.~d.~S. Cicero, Y.~Mo, X.~Bing, Z.~Bowen, and
  B.~Yoshua, ``A structured self-attentive sentence embedding,'' \emph{arXiv
  preprint arXiv:1703.03130}, 2017.

\bibitem{frasconi1998graph}
F.~Paolo, G.~Marco, and S.~Alessandro, ``A general framework for adaptive
  processing of data structures,'' \emph{IEEE transactions on Neural Networks},
  1998.

\bibitem{sperduti1997graph}
A.~Sperduti and A.~Starita, ``Supervised neural networks for the classification
  of structures,'' \emph{IEEE Transactions on Neural Networks}, 1997.

\bibitem{gori2005graph}
G.~Marco, M.~Gabriele, and F.~Scarselli, ``A new model for learning in graph
  domains,'' in \emph{IJCNN}, 2005.

\bibitem{scarselli2009graph}
S.~Franco, G.~Marco, T.~Ah~Chung, H.~Markus, and M.~Gabriele, ``The graph
  neural network,'' \emph{IEEE Transactions on Neural Networks}, 2009.

\bibitem{velickovic2017graphAN}
P.~Velickovic, G.~Cucurull, A.~Casanova, A.~Romero, P.~Li{\`o}, and Y.~Bengio,
  ``Graph attention networks,'' \emph{ArXiv}, vol. abs/1710.10903, 2017.

\bibitem{cimpoi2014dtd}
M.~Cimpoi, S.~Maji, I.~Kokkinos, S.~Mohamed, and A.~Vedaldi, ``Describing
  textures in the wild,'' in \emph{CVPR}, 2014.

\bibitem{bell2015minc}
S.~Bell, P.~Upchurch, N.~Snavely, and K.~Bala, ``Material recognition in the
  wild with the materials in context database,'' in \emph{CVPR}, 2015.

\bibitem{paszke2017automatic}
A.~Paszke, S.~Gross, S.~Chintala, G.~Chanan, E.~Yang, Z.~DeVito, Z.~Lin,
  A.~Desmaison, L.~Antiga, and A.~Lerer, ``Automatic differentiation in
  {PyTorch},'' in \emph{NeurIPS Autodiff Workshop}, 2017.

\bibitem{imagenet_cvpr09}
J.~Deng, W.~Dong, R.~Socher, L.-J. Li, K.~Li, and L.~Fei-Fei, ``Imagenet: A
  large-scale hierarchical image database,'' in \emph{CVPR}, 2009.

\bibitem{tenenbaum1997bilinear}
J.~B. Tenenbaum and W.~T. Freeman, ``Separating style and content,'' in
  \emph{Advances in neural information processing systems}, 1997.

\bibitem{lin2015bilinear}
T.-Y. Lin, A.~RoyChowdhury, and S.~Maji, ``Bilinear cnn models for fine-grained
  visual recognition,'' in \emph{ICCV}, 2015.

\bibitem{cimpoi2015fvcnn}
M.~Cimpoi, S.~Maji, and A.~Vedaldi, ``Deep filter banks for texture recognition
  and segmentation,'' in \emph{CVPR}, 2015.

\end{thebibliography}
%



\end{document}